\documentclass[11pt]{article}

\usepackage[preprint]{acl}

\usepackage{times}
\usepackage{latexsym}

\usepackage[T1]{fontenc}

\usepackage[utf8]{inputenc}

\usepackage{microtype}

\usepackage{booktabs}

\usepackage{amsmath}
\usepackage{amssymb}

\usepackage{inconsolata}

\usepackage{enumitem}

\usepackage{graphicx}

\usepackage{tabularx}   
\usepackage{makecell}   

\usepackage{subcaption}

\usepackage{fancyvrb}
\usepackage{color}
\usepackage[many]{tcolorbox}
\usepackage{amsmath}
\usepackage{amsfonts}
\definecolor{green_first}{RGB}{168, 209, 176}   
\definecolor{green_second}{RGB}{200, 235, 200}  
\definecolor{green_third}{RGB}{235, 255, 235}   
\definecolor{light_green_table}{RGB}{220, 255, 220}  
\definecolor{light_purple_table}{RGB}{235, 225, 255}  
\definecolor{light_green}{rgb}{0.569, 0.800, 0.459}
\definecolor{blue_dist}{rgb}{0.192,0.443,0.651}
\definecolor{orange_dist}{rgb}{0.812,0.545,0.239}
\definecolor{yellow_dist}{rgb}{0.918,0.804,0.463}
\definecolor{website}{rgb}{0.9333333333333333, 0.10980392156862745, 0.592156862745098} 
\definecolor{pm_rowcolor}{rgb}{0.85, 0.90, 0.84} 
\definecolor{medium_gray}{RGB}{150, 150, 150}  
\definecolor{medium_purple}{RGB}{150, 120, 200}  
\newtcolorbox{promptbox}[2][Prompt]{
    colback=black!5!white,        
    arc=5pt,                      
    boxrule=0.5pt,                
    fonttitle=\bfseries,          
    title=#1,                     
    before upper={\small},        
    fontupper=\fontfamily{ptm}\selectfont, 
    colframe=#2,                  
    left=3pt,                     
    right=3pt,                    
    top=3pt,                      
    bottom=3pt,                   
    boxsep=3pt,                   
    toptitle=1pt,                 
    bottomtitle=1pt,              
    lefttitle=1pt,                
    righttitle=1pt,               
}
\usepackage{fontawesome}
\usepackage{multirow}
\usepackage{subcaption}
\usepackage{wrapfig}
\usepackage{algorithm}
\usepackage{algpseudocode}
\usepackage{amsmath}
\usepackage{float}
\usepackage{verbatim}

\usepackage{algpseudocode}

\title{Distributional Clarity: The Hidden Driver of RL-Friendliness in Large Language Models}

\author{
Shaoning Sun$^{1}$\thanks{\ \ Equal contribution.}\thanks{\ \ Work done during internship at Baidu.},
Mingzhu Cai$^{2}$\footnotemark[1],
Huang He$^{2}$,
Bingjin Chen$^{2}$,
\\
\textbf{Siqi Bao}$^{2}$\thanks{\raggedright\ \ Corresponding authors: \texttt{yang.yujiu@sz.tsinghua.edu.cn, baosiqi@baidu.com}.},
\textbf{Yujiu Yang}$^{1}$\footnotemark[3],
\textbf{Hua Wu}$^{2}$,
\textbf{Haifeng Wang}$^{2}$
\\
$^1$Tsinghua Shenzhen International Graduate School, Tsinghua University
\\
$^2$Baidu Inc.
}

\begin{document}
\maketitle
\begin{abstract}

Language model families exhibit striking disparity in their capacity to benefit from reinforcement learning: under identical training, models like Qwen achieve substantial gains, while others like Llama yield limited improvements. Complementing data-centric approaches, we reveal that this disparity reflects a hidden structural property: \textbf{distributional clarity} in probability space. Through a three-stage analysis—from phenomenon to mechanism to interpretation—we uncover that RL-friendly models exhibit intra-class compactness and inter-class separation in their probability assignments to correct vs. incorrect responses. We quantify this clarity using the \textbf{Silhouette Coefficient} ($S$) and demonstrate that (1) high $S$ correlates strongly with RL performance; (2) low $S$ is associated with severe logic errors and reasoning instability. To confirm this property, we introduce a Silhouette-Aware Reweighting strategy that prioritizes low-$S$ samples during training. Experiments across six mathematical benchmarks show consistent improvements across all model families, with gains up to 5.9 points on AIME24. Our work establishes distributional clarity as a fundamental, trainable property underlying RL-Friendliness.

\end{abstract}

\section{Introduction}
Reinforcement learning with verifiable rewards (RLVR) has become the dominant approach for enhancing LLM reasoning \citep{deepseekr1,openreasoner,rlpr,beyond,surprising}. However, beneath this success lies a puzzling asymmetry: when trained with identical RLVR pipelines such as GRPO \citep{deepseekmath} and DAPO \citep{dapo}, Qwen models \citep{qwen2.5} consistently achieve substantial gains in mathematical reasoning, while Llama \citep{llama3} yields only limited improvements \citep{understanding,cognitive,simplerl}. This disparity reflects differences in \textit{RL-Friendliness}—the capacity of a model to benefit from reinforcement learning.

\textbf{Why do foundation models differ in their capacity to benefit from RL training?} Prior work has approached this question from a data-centric perspective. \citet{cognitive} identified specific reasoning patterns as critical differentiators, demonstrating that fine-tuning with data containing these patterns can partially bridge the gap during subsequent RL training. Similarly, OctoThinker \citep{octothinker} showed that exposing Llama to high-quality mathematical corpora during mid-training better prepares it for reinforcement learning. In parallel to these data-centric approaches, we examine an orthogonal aspect: the intrinsic generation properties of different model families, aiming to uncover the key drivers of RL-Friendliness.

\begin{figure}[t]
    \centering
    \includegraphics[width=0.95\linewidth]{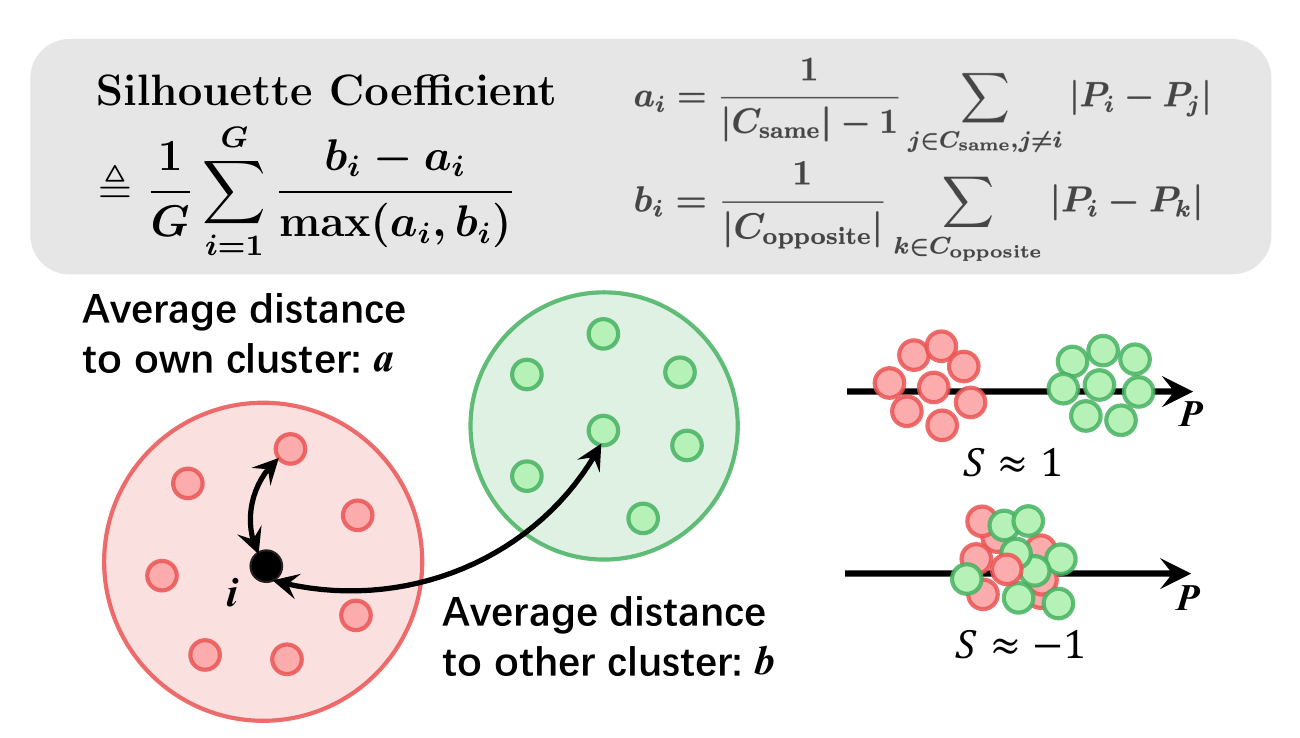}
    \vspace{-0.5em}
    \caption{Schematic illustration of the Silhouette Coefficient ($S$). We adapt this metric to quantify distributional clarity. High $S$ values represent ideal landscapes with compact and separated clusters, while low values indicate overlapping distributions.}
    \label{fig:silhouette_schema}
    \vspace{-1.2em}
\end{figure}

We introduce a three-stage analysis framework that systematically dissects RL-Friendliness \textit{from phenomenon to mechanism to interpretation}. At the phenomenological level, we find that while RL-friendly and less RL-friendly models solve largely overlapping problem sets, RL-friendly models achieve consistently higher pass rates on these shared problems, implying superior probability assignment to correct solutions. At the mechanistic stage, we identify a key distributional difference on probability: RL-friendly models exhibit \textbf{intra-class compactness and inter-class separation}-probability scores for correct and incorrect responses cluster densely within their groups yet remain significantly separated. We introduce the \textbf{Silhouette Coefficient ($S$)} \citep{probabilistic} to quantify this distributional clarity and observe a strong positive correlation between $S$ and pass rates.

At the interpretative level, we show that distributional clarity governs reasoning quality at the semantic level by examining error attribution and solution stability. RL-friendly models produce predominantly low-severity errors (calculation issues) and exhibit consistent reasoning paths, whereas less RL-friendly models generate high-severity errors (logic flaws) and show unstable reasoning. Crucially, samples with low $S$ are disproportionately associated with severe errors and instability, while high-$S$ samples show stable reasoning and minor errors. This reveals that distributional clarity constitutes a key factor in effective RL optimization.

To validate the critical role of distributional clarity in RL-Friendliness, we introduce a \textit{Silhouette-Aware Reweighting} strategy that prioritizes low-$S$ samples during training. By targeting samples with poor distributional clarity, we force the model to improve its worst-performing areas while maintaining its strengths. Experiments on six mathematical reasoning benchmarks demonstrate consistent improvements for all model families, particularly on challenging datasets like AIME24, where gains range from 1.8 to 5.9 points across model families. These gains across diverse models validate our analysis and demonstrate that addressing distributional clarity is key to enhance RL-Friendliness.

In summary, our contributions are as follows:
\begin{enumerate}[leftmargin=13pt, topsep=0pt, itemsep=0pt, parsep=3pt]
    \item \textbf{A three-stage diagnostic framework.} We demonstrate that RL-Friendliness reflects not merely \textit{what} models can solve, but \textit{how reliably} they solve it. Through systematic analysis from phenomenon to mechanism to interpretation, we identify distributional clarity—specifically intra-class compactness and inter-class separation in the probability assignments to correct vs.\ incorrect responses—as the fundamental structural property governing RL-Friendliness.
    
    \item \textbf{A mechanistic understanding.} We introduce the Silhouette Coefficient ($S$) to quantify distributional clarity and show that it serves as a unified metric bridging structure and behavior: it simultaneously reflects performance, correlates with error severity, and tracks reasoning stability. This establishes \textit{how} distributional clarity governs RL-Friendliness: high $S$ ensures stable training dynamics aligned with optimization goals, while low $S$ manifests as severe logical errors and unstable reasoning—impeding the reinforcement of reliable behaviors.

    \item \textbf{A practical intervention.} We introduce a Silhouette-Aware Reweighting strategy to validate our analysis, which prioritizes low-$S$ samples during training. Experiments across six benchmarks yield consistent gains for all model families, confirming that distributional clarity is not merely explanatory but trainable—transforming structural diagnosis into practical enhancement.
\end{enumerate}

\section{Anatomy of RL-Friendliness: A Three-Stage Analysis}
\label{sec:three_level}

In this section, we conduct a systematic analysis to uncover the underlying factors contributing to the disparity in RL-friendliness across different model families. We apply our three-stage framework, progressing from phenomenon to mechanism to interpretation. At the \textit{phenomenological level}, we examine performance to quantify outcome disparities on shared problem sets. At the \textit{mechanistic level}, we analyze probability distributions to identify the structural characteristics that drive these performance differences. At the \textit{interpretative level}, we assess reasoning behaviors to provide semantic interpretations of these distributional patterns in terms of error severity and solution stability.

\subsection{Phenomenon: The Performance Disparity}
\label{sec:outcome_level}

To investigate RL-Friendliness disparities, we begin by quantifying model performance through \textit{pass rates}—the empirical probability of generating correct responses. For each query $q$, we sample $K$ independent responses and compute the pass rate $\rho(q)$—the fraction verified as correct. Higher $\rho(q)$ indicates greater probability mass assigned to correct reasoning paths.

\begin{figure}[t]
    \centering
    \includegraphics[width=0.98\linewidth]{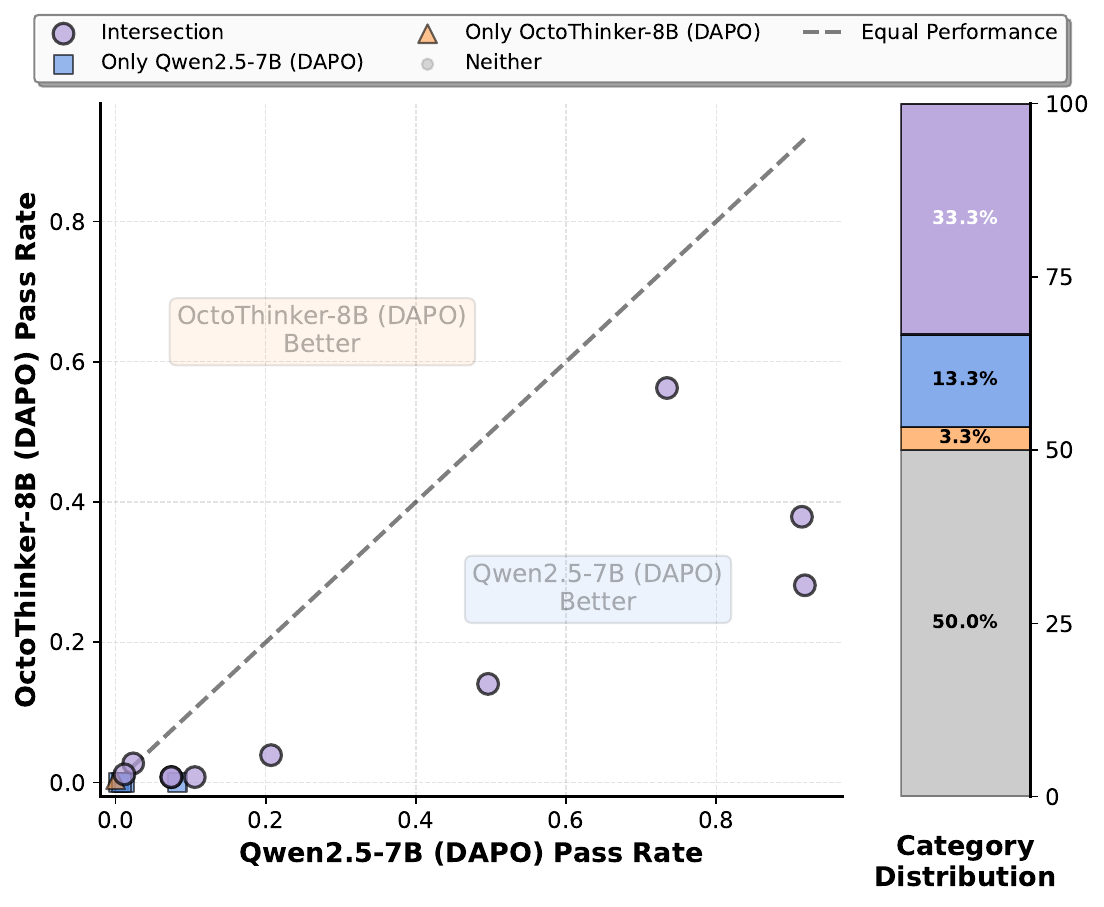}
    \vspace{-0.5em}
    \caption{Per-problem pass rate comparison between Qwen2.5-7B (DAPO) and OctoThinker-8B (DAPO) on AIME 2024. Each point represents a query. Points below the diagonal indicate Qwen achieves higher pass rates. A similar distribution pattern is observed on MATH-500 (see Figure \ref{fig:pass_rate_comparison_math500}).}
    \label{fig:pass_rate_comparison}
    \vspace{-1.2em}
\end{figure}

We conducted a comparative analysis using Qwen2.5-7B \citep{qwen2.5} and OctoThinker-8B-Hybrid-Base \citep{octothinker}, both trained using the DAPO algorithm under identical settings. Figure \ref{fig:pass_rate_comparison} visualizes the per-query pass rates on AIME 2024 ($K=256$). We focus on the intersection of solvable problems—queries where both models achieve $\rho > 0$.

Figure \ref{fig:pass_rate_comparison} shows substantial overlap in the fundamental capabilities of the two models. Specifically, the intersection set accounts for approximately 71.4\% of the total solvable problems for Qwen and 90.9\% for OctoThinker. However, within this intersection, the behavior diverges markedly. Most data points fall below the diagonal ($y=x$), indicating that Qwen consistently achieves higher pass rates than OctoThinker on the \text{same} problems.

This asymmetry reveals a critical insight: \textbf{RL-Friendliness is not merely about what a model can solve, but also how reliably it solves it.} Less RL-friendly models fail to assign sufficient probability mass to the correct solutions, while RL-friendly models maintain high reliability in generating correct responses for shared solvable instances.

\subsection{Mechanism: Compactness and Separation}
\label{sec:distributional_level}

The performance disparity observed in Section~\ref{sec:outcome_level}, we hypothesize, arises from fundamental differences in how models distribute confidence over correct and incorrect responses. To investigate the mechanism driving this behavior, we quantify response confidence as the geometric mean of token-level probabilities, computing a length-normalized sequence score $P(o|q)$. Given a query $q$ and a generated response $o = (y_1, y_2, \dots, y_L)$, we calculate:
\begin{equation}
\setlength{\abovedisplayskip}{5pt}
\setlength{\belowdisplayskip}{5pt}
    P(o|q) = \left( \prod_{i=1}^{L} P(y_i | q, y_{<i}) \right)^{\frac{1}{L}}
\end{equation}

\begin{figure}[t]
    \centering
    \includegraphics[width=0.95\linewidth]{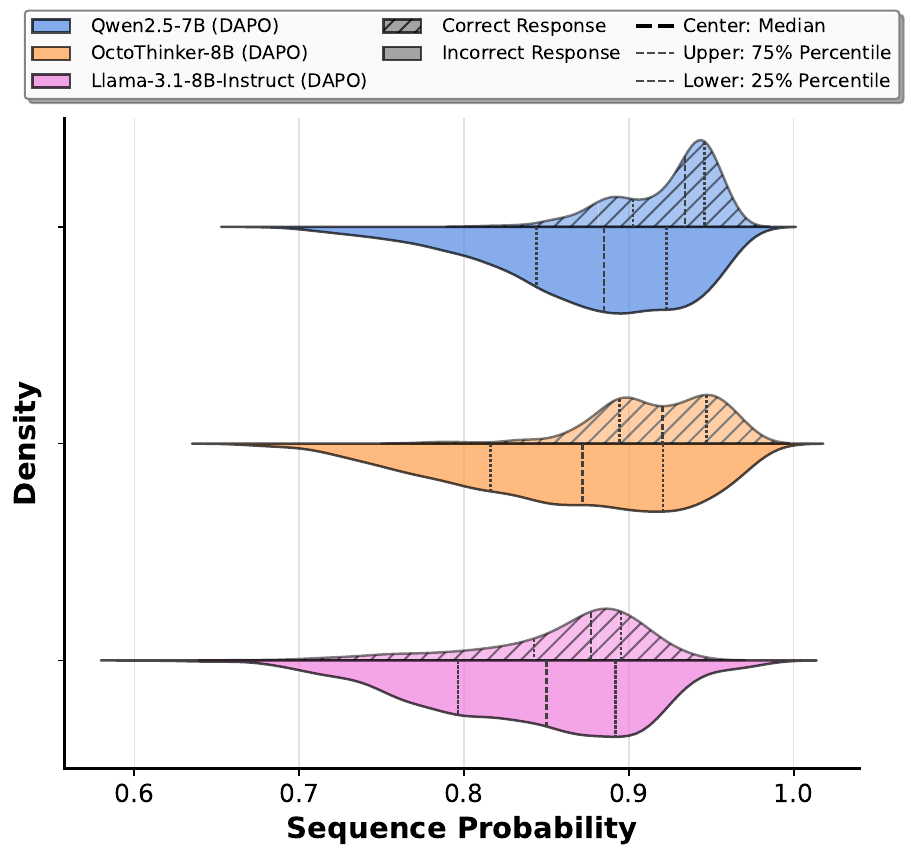}
    \vspace{-0.5em}
    \caption{Probability Distributions: Kernel density estimates of sequence probabilities for correct (top) and incorrect (bottom) responses. Qwen exhibits clear separation, whereas Llama and OctoThinker show significant overlap.}
    \label{fig:dist_landscape}
    \vspace{-1.2em}
\end{figure}
Figure~\ref{fig:dist_landscape} visualizes the distribution of sequence-level probabilities for correct versus incorrect responses for Qwen2.5-7B, Llama-3.1-8B, and OctoThinker-8B (all DAPO-trained) on AIME 2024. These distributions reveal striking structural differences across models. Most notably, Qwen's distribution exhibits two distinct characteristics:
\begin{itemize}[leftmargin=13pt, topsep=0pt, itemsep=0pt]
    \item \textbf{intra-class compactness}—probability scores cluster tightly within correct and incorrect groups
    \item \textbf{inter-class separation}—a substantial margin exists between the correct and incorrect clusters
\end{itemize}
In contrast, Llama and OctoThinker display significant overlap: probabilities of incorrect responses frequently match or exceed those of correct ones, creating ambiguous decision boundaries.

To rigorously quantify this structural property, we introduce the \textbf{Silhouette Coefficient} ($S$), adapted from cluster analysis to our one-dimensional probability distributions. For a given query, let $\mathcal{O}$ denote the set of all generated response probabilities, which is partitioned into two clusters: correct responses and incorrect responses. For each response score $P_i \in \mathcal{O}$, let $C_{\text{same}}$ be the cluster containing $P_i$ and $C_{\text{opposite}}$ be the complementary cluster. We compute the average intra-cluster distance $a_i$ and the average inter-cluster distance $b_i$ as follows:
{
\setlength{\abovedisplayskip}{5pt}
\setlength{\belowdisplayskip}{5pt}
\begin{align}
    a_i &= \frac{1}{|C_{\text{same}}|-1} \sum_{P_j \in C_{\text{same}}, j \neq i} |P_i - P_j| \\
    b_i &= \frac{1}{|C_{\text{opposite}}|} \sum_{P_k \in C_{\text{opposite}}} |P_i - P_k|
\end{align}
}
Based on these distances, the Silhouette value $s_i$ for the individual sample is defined as:
\begin{equation}
\setlength{\abovedisplayskip}{5pt}
\setlength{\belowdisplayskip}{5pt}
    s_i = \frac{b_i - a_i}{\max(a_i, b_i)}
\end{equation}
The final query-level coefficient $S \in [-1, 1]$ is obtained by averaging $s_i$ across all samples (i.e., $S = \frac{1}{|\mathcal{O}|}\sum_{P_i \in \mathcal{O}} s_i$). As illustrated in Figure \ref{fig:silhouette_schema}, this metric captures total distributional quality: values approaching 1 indicate ideal structure—compact clusters with clear separation—while negative values reveal overlap between correct and incorrect groups.

These two properties—compactness and separation—govern RL-Friendliness through different mechanisms. We provide theoretical grounding in Appendix~\ref{sec:gradient_variance}, showing that gradient variance in GRPO-style algorithms is predominantly determined by the probability distributions of correct and incorrect responses. When these distributions exhibit high compactness, the gradient signal remains stable, ensuring consistent parameter updates. Moreover, clear separation between correct and incorrect clusters directly aligns with the RL training objective, which is to increase the probability of correct responses while decreasing that of incorrect ones. \textbf{High $S$ thus indicates both stable training dynamics and alignment with the optimization goal.}

\begin{figure}[t]
    \centering
    \includegraphics[width=0.9\linewidth]{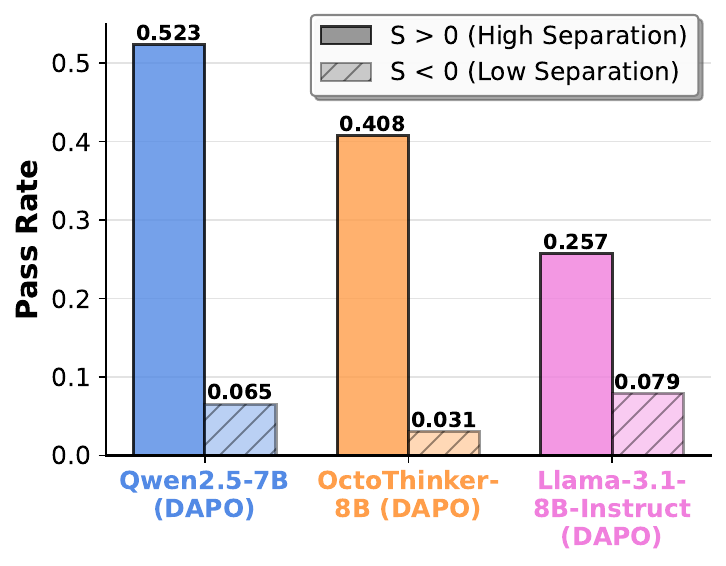}
    \vspace{-0.5em}
    \caption{Impact of distributional structure on pass rates. Queries with positive $S$ values achieve significantly higher performance across all models.}
    \label{fig:passrate_by_silhouette}
    \vspace{-1.2em}
\end{figure}

\begin{figure*}[!t]
    \centering
    \begin{subfigure}[b]{0.54\textwidth}
        \centering
        \includegraphics[width=\linewidth]{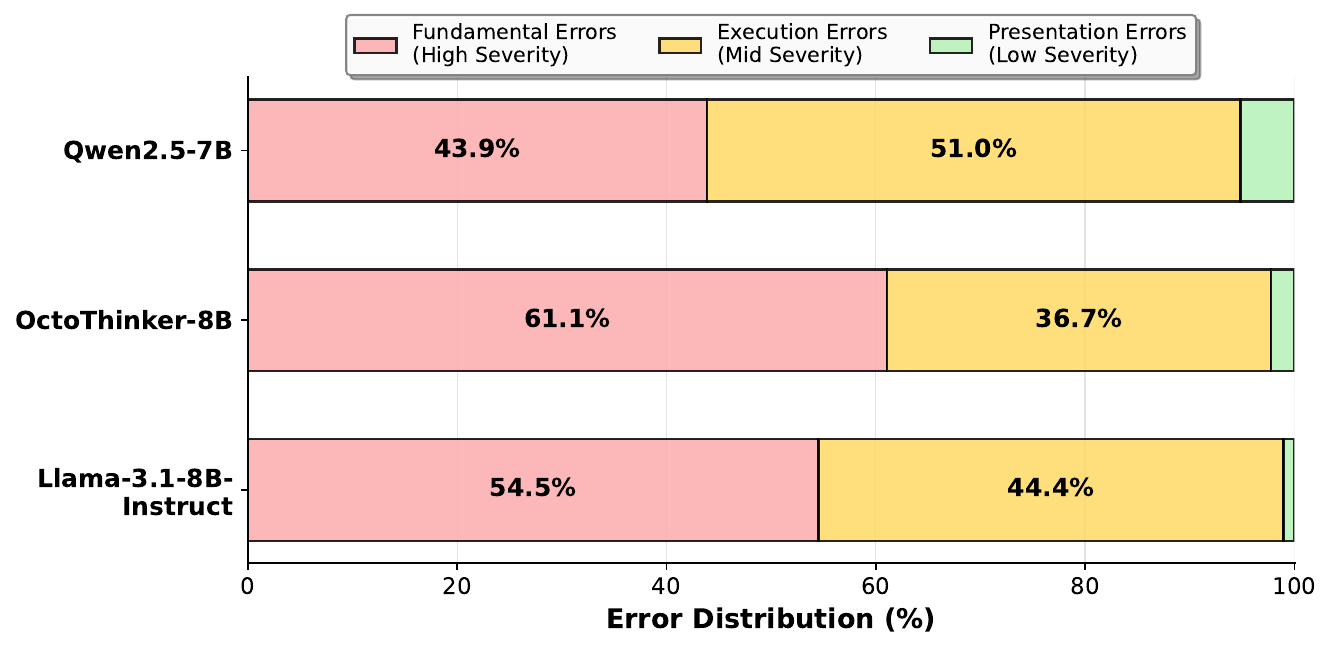}
        \caption{Distribution of Error Severity}
        \label{fig:error_severity_dist}
    \end{subfigure}
    \hfill
    \begin{subfigure}[b]{0.44\textwidth}
        \centering
        \includegraphics[width=\linewidth]{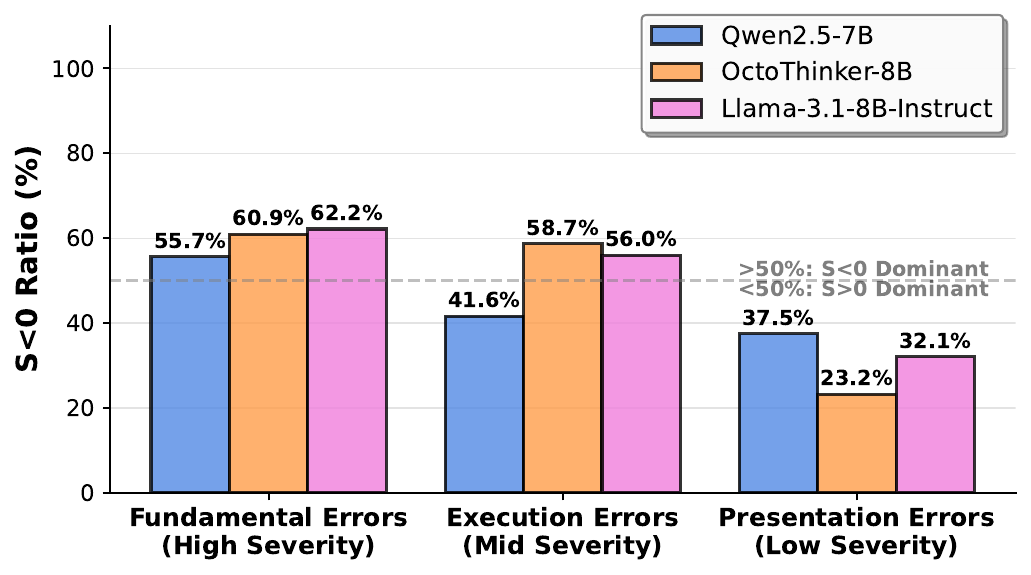}
        \caption{Error Severity and Silhouette Coefficient Relationship}
        \label{fig:error_silhouette_corr}
    \end{subfigure}
    \vspace{-0.5em}
    \caption{Error attribution analysis on MATH-500. (a) Proportion of High (Fundamental), Mid (Execution), and Low (Presentation) severity errors across models. (b) Percentage of responses with $S < 0$ (poor distributional clarity) within each error category. High-severity errors are strongly correlated with low distributional clarity.}
    \label{fig:error_analysis}
    \vspace{-1.2em}
\end{figure*}

Empirically, \textbf{this distributional clarity positively correlates with performance}. Figure~\ref{fig:passrate_by_silhouette} compares pass rates across queries partitioned by $S$ into high-clarity ($S > 0$) and low-clarity ($S < 0$) groups.\footnote{We exclude queries where all responses are correct or all incorrect, as $S$ is undefined in these cases.} The gaps are substantial: for Qwen, high-$S$ queries achieve 52.3\% pass rates versus 6.5\% for low-$S$ queries—an 8$\times$ difference. OctoThinker exhibits an even more dramatic pattern: 40.8\% versus 3.1\%, a 13$\times$ gap. This validates $S$ as a reliable indicator of model performance and RL training potential.

\subsection{Interpretation: Error Severity and Solution Stability}
\label{sec:behavioral_level}

To understand \textit{what} distributional clarity corresponds to semantically, we conduct a fine-grained behavioral analysis on MATH-500 along two dimensions: (i) the nature of generated errors and (ii) the stability of the reasoning strategies employed.

\noindent \textbf{Error Severity Attribution.}
We first investigate whether the incorrect responses from different models stem from similar types of failures. Using an LLM-as-a-judge approach\footnote{Detailed error taxonomy definitions and evaluation prompts are provided in Appendix~\ref{sec:appendix_error}.}, we construct a hierarchical error taxonomy and classify failures into three severity levels: \textit{High Severity (Fundamental)} involving core logic or knowledge flaws; \textit{Mid Severity (Execution)} involving calculation errors; and \textit{Low Severity (Presentation)} involving formatting issues. Figure \ref{fig:error_severity_dist} shows that Qwen exhibits a significantly healthier error profile, with a lower proportion of high-severity errors compared to OctoThinker and Llama.

Crucially, we find a clear association: \textbf{high-severity errors are disproportionately concentrated in responses with poor distributional clarity}. Figure~\ref{fig:error_silhouette_corr} reports, for each error category, the fraction of responses with negative Silhouette Coefficients ($S<0$). For Llama, 62.2\% of fundamental errors have $S<0$, indicating weak separation between correct and incorrect responses when the model makes logic-breaking mistakes. In contrast, low-severity errors are more likely to retain $S>0$, suggesting that the probability distribution correctly identifies promising approaches.

\noindent \textbf{Solution Stability Analysis.}
Beyond error analysis, we measure the stability of correct reasoning by computing the proportion of distinct solution methods among all correct responses for each query. Two solutions are deemed to share the \text{same} method if they utilize identical key theorems, formulas, or logical strategies, ignoring variations in variable naming or verbosity. We cluster correct responses using an automated judge. Clustering details and prompts are in Appendix~\ref{sec:appendix_solution_stability}.

\begin{figure*}[t]
    \centering
    \begin{subfigure}[b]{0.48\textwidth}
        \centering
        \includegraphics[width=\linewidth]{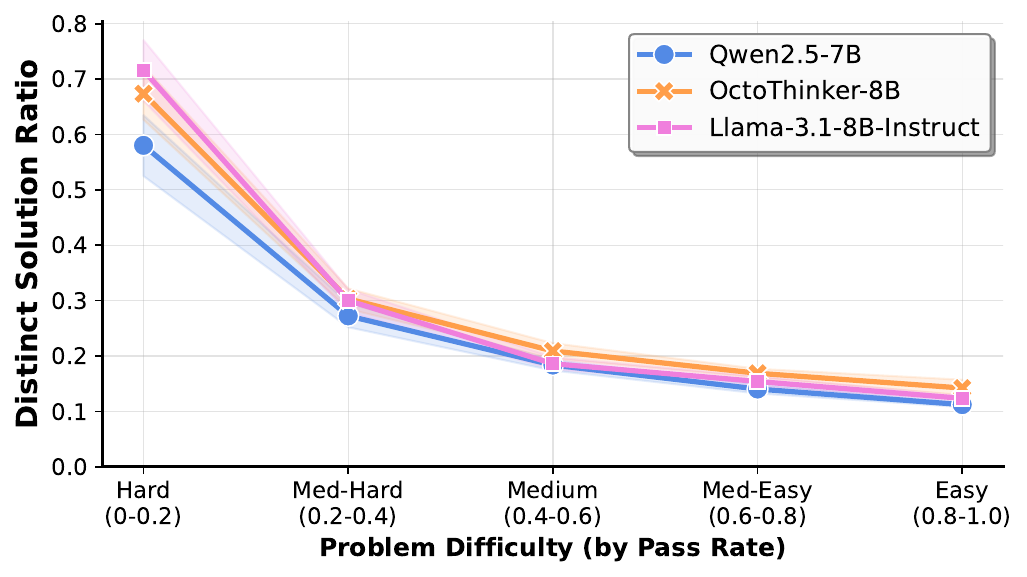}
        \caption{Distinct Solution Ratio vs. Problem Difficulty}
        \label{fig:method_density_diff}
    \end{subfigure}
    \hfill
    \begin{subfigure}[b]{0.48\textwidth}
        \centering
        \includegraphics[width=\linewidth]{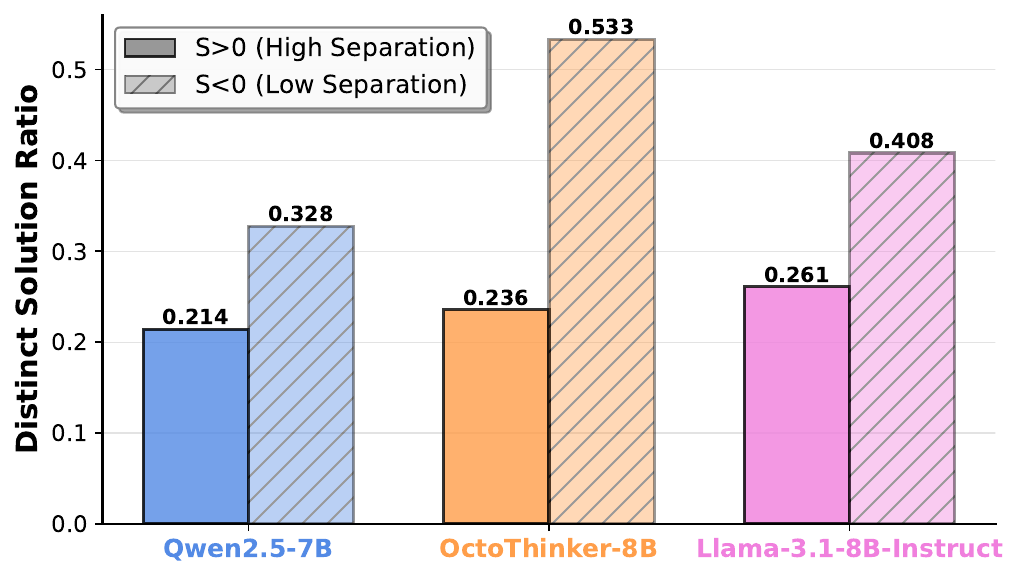}
        \caption{Distinct Solution Ratio by Distributional Clarity}
        \label{fig:diversity_clarity}
    \end{subfigure}
    \vspace{-0.5em}
    \caption{Solution stability analysis. (a) Average distinct solution ratio across problem difficulty tiers. Lower ratio implies greater consistency. (b) Solution ratio for queries grouped by Silhouette Coefficient. High clarity ($S \ge 0$) correlates with more stable, consistent reasoning patterns.}
    \label{fig:stability_analysis}
    \vspace{-1.2em}
\end{figure*}

Distinct solutions ratio differs systematically between RL-friendly and less RL-friendly models. As Figure~\ref{fig:method_density_diff} shows, Qwen exhibits consistently low ratio, particularly on hard problems, indicating stable convergence to effective strategies. In contrast, OctoThinker and Llama show significantly more distinct solutions. Closer inspection reveals that high ratio in less RL-friendly models reflects reasoning instability rather than genuine methodological breadth: these models frequently reach correct answers through spurious correctness and hallucinated steps that register as distinct methods, inflating the distinct ratio metric.

We demonstrate that this instability is directly tied to distributional clarity. Figure \ref{fig:diversity_clarity} compares distinct solution ratio for queries with $S \ge 0$ versus $S < 0$. Across all models, high-clarity queries exhibit significantly lower ratio, meaning the model consistently reproduces the similar valid reasoning logic. Conversely, low-clarity queries produce fragmented reasoning paths. This establishes the causal chain: \textbf{poor distributional clarity ($S < 0$) generates reasoning instability (high ratio), which fundamentally undermines RL training by preventing reliable identification of behaviors to reinforce}. Conversely, high $S$ enables stable reasoning policies that RL can effectively reinforce.

\section{Empirical Validation via Silhouette-Aware Reweighting}

To validate that distributional clarity drives RL-Friendliness, we propose Silhouette-Aware Reweighting—a strategy that modulates the training signal to prioritize queries with poor distributional clarity. We adopt DAPO \citep{dapo} as our training backbone.

\noindent \textbf{Standard Formulation.}
Formally, given a query $q$ and a group of $G$ outputs $\{o_i\}_{i=1}^G$ with rewards $R_i$, the standard advantage $\hat{A}_{i,t}$ in DAPO (and GRPO) is computed via group normalization:
\begin{equation}
\setlength{\abovedisplayskip}{5pt}
\setlength{\belowdisplayskip}{5pt}
    \hat{A}_{i,t} = \frac{R_i - \mu(\{R_j\}_{j=1}^G)}{\sigma(\{R_j\}_{j=1}^G)}
    \label{eq:advantage_norm}
\end{equation}
where $\mu$ and $\sigma$ denote the mean and standard deviation of the group rewards.

\begin{figure}[h]
    \centering
    \includegraphics[width=0.8\linewidth]{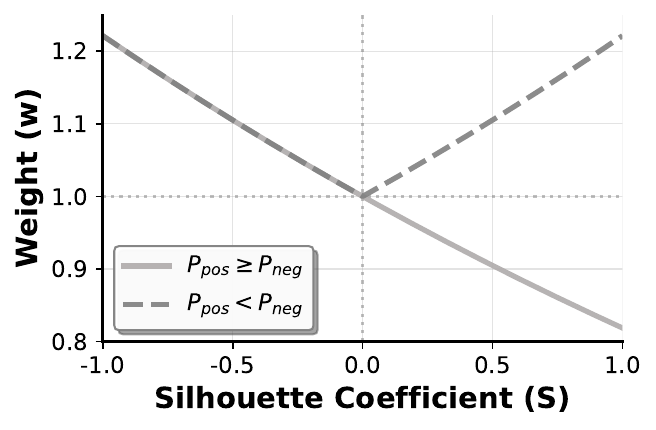}
    \caption{Silhouette-Aware Reweighting function $w(S) = \exp(-\beta \cdot S)$ with $\beta = 0.2$.}
    \label{fig:weight_curve}
    \vspace{-1.2em}
\end{figure}

\noindent \textbf{Silhouette-Aware Reweighting.}
To explicitly force the model to focus on samples exhibiting distributional ambiguity, we introduce a weighted advantage $\tilde{A}_{i,t}$ defined as:
\begin{equation}
\setlength{\abovedisplayskip}{5pt}
\setlength{\belowdisplayskip}{5pt}
    \tilde{A}_{i,t} = \hat{A}_{i,t} \cdot w(q)
\end{equation}
where $w(q)$ is a query-specific weight derived from the distributional properties of the model outputs.

To handle cases where the model assigns higher probabilities to incorrect responses, we introduce a rectified Silhouette metric. Let $P_{pos}$ and $P_{neg}$ denote the average sequence-level probabilities for correct and incorrect response groups. When $P_{pos} < P_{neg}$, simply optimizing for high $S$ values might inadvertently push the distributions further apart in the wrong direction. To mitigate this, we define:
\begin{equation}
\setlength{\abovedisplayskip}{5pt}
\setlength{\belowdisplayskip}{5pt}
    S' = 
    \begin{cases} 
    S & \text{if } P_{pos} \ge P_{neg} \\
    -|S| & \text{if } P_{pos} < P_{neg}
    \end{cases}
\end{equation}
This rectified metric becomes strictly negative when the distribution is inverted, ensuring the reweighting emphasizes reversing the probability mass back to the correct direction.

Based on this rectified metric, we define the reweighting factor $w(q)$ as follows:
\begin{equation}
\label{eq:reweight}
\setlength{\abovedisplayskip}{5pt}
\setlength{\belowdisplayskip}{5pt}
    w(q) = \exp(-\beta \cdot S')
\end{equation}
where $\beta > 0$ controls the sensitivity of the reweighting. Figure \ref{fig:weight_curve} illustrates this function assigns larger weights ($w > 1$) to queries with negative $S'$, amplifying the gradient signal for samples with poor or inverted distributional clarity. Conversely, for queries with high positive $S'$ where the distribution is already well-structured, the weight decays ($w < 1$), preventing the model from overfitting to easy samples.

\section{Experiments}

\subsection{Experimental Setup}
\label{sec:experimental_setup}
\noindent \textbf{Models and Datasets.}
We validate our approach on three backbone models: Qwen2.5-7B \citep{qwen2.5}, OctoThinker-8B-Hybrid-Base \citep{octothinker}, and Llama-3.1-8B-Instruct \citep{llama3}. OctoThinker-8B-Hybrid-Base is a model derived from Llama-3.1-8B via mid-training on reasoning corpora. For the Llama family, we select the Instruct version because the Base model exhibits insufficient instruction-following capabilities for direct RLVR. Regarding training data, we utilize the DAPO-Math-17k dataset \citep{dapo} for both Qwen and OctoThinker. For Llama, we utilize the MATH dataset \citep{math} following prior studies \citep{simplerl,does} due to its initial reasoning performance.

\noindent \textbf{Evaluation Benchmarks.}
We select six widely recognized benchmarks in the mathematical reasoning domain for testing: AIME 2024, AIME 2025, MATH-500 \citep{math}, AMC, Minerva \citep{minerva}, and OlympiadBench \citep{olympiad}.

\noindent \textbf{Implementation Details.}
The reweighting hyperparameter $\beta$ is set to 0.2. Additional details regarding the training configurations and evaluation setting are provided in Appendix \ref{sec:appendix_experimental_setup}.

\subsection{Main Results}
\label{sec:main_results}

\begin{table*}[t]
\centering
\small
\setlength{\tabcolsep}{4pt} 
\resizebox{0.9\textwidth}{!}{
\begin{tabular}{lccccccc}
\toprule
\multirow{2}{*}{\textbf{Model}} & \textbf{AIME24} & \textbf{AIME25} & \textbf{MATH500} & \textbf{AMC23} & \textbf{Minerva} & \textbf{Olympiad} & \multirow{2}{*}{\textbf{Average}} \\
 & \textit{avg@256} & \textit{avg@256} & \textit{avg@32} & \textit{avg@16} & \textit{avg@16} & \textit{avg@16} & \\
\midrule
\multicolumn{8}{l}{\textit{\textbf{Qwen Family}}} \\ 
Qwen2.5-7B & 5.5 & 2.6 & 50.7 & 30.5 & 19.4 & 23.1 & 22.0 \\
+ DAPO & 12.2 & 11.8 & 79.7 & \textbf{70.2} & \textbf{36.4} & 42.4 & 42.1 \\
+ DAPO-Silhouette & \textbf{18.1} & \textbf{12.0} & \textbf{80.4} & \textbf{70.2} & 34.3 & \textbf{43.2} & \textbf{43.0} \\
\midrule
\multicolumn{8}{l}{\textit{\textbf{OctoThinker Family}}} \\
OctoThinker-8B & 1.5 & 0.7 & 39.0 & 18.0 & 13.3 & 13.2 & 14.3 \\
+ DAPO & 4.9 & 2.1 & 59.4 & 46.9 & 27.6 & 25.8 & 27.8 \\
+ DAPO-Silhouette & \textbf{8.2} & \textbf{5.0} & \textbf{62.6} & \textbf{47.2} & \textbf{29.6} & \textbf{27.8} & \textbf{30.1} \\
\midrule
\multicolumn{8}{l}{\textit{\textbf{Llama Family}}} \\
Llama-3.1-8B-Instruct & 3.7 & 0.4 & 46.8 & 24.7 & 21.7 & 15.4 & 18.8 \\
+ DAPO & 6.1 & \textbf{0.5} & 51.4 & 25.2 & 25.1 & \textbf{19.4} & 21.3 \\
+ DAPO-Silhouette & \textbf{7.9} & 0.4 & \textbf{53.1} & \textbf{27.7} & \textbf{25.4} & \textbf{19.4} & \textbf{22.3} \\
\bottomrule
\end{tabular}
}
\vspace{-0.5em}
\caption{Main results across six mathematical reasoning benchmarks. Our Silhouette-aware strategy consistently outperforms the standard DAPO baseline, with particularly significant gains for less RL-friendly models (OctoThinker and Llama) on challenging benchmarks like AIME24.}
\label{tab:main_results}
\vspace{-1.2em}
\end{table*}

Table \ref{tab:main_results} presents the performance comparison between DAPO and our Silhouette-aware variant, providing strong empirical validation for our analysis.

\noindent \textbf{Unlocking Potential in Less RL-Friendly Models.}
Our method outperforms the baseline across all models, with the most significant average gains observed in less RL-friendly families.
For OctoThinker, the approach nearly doubles the pass rates on AIME 2024 (4.9\% $\rightarrow$ 8.2\%) and AIME 2025 (2.1\% $\rightarrow$ 5.0\%). Llama also achieves notable improvements on AIME 2024 and Math500. These results confirm that explicitly targeting samples with poor distributional clarity effectively alleviates the optimization barrier for these models.

\noindent \textbf{Enhancing Strong Models.}
Even for the already RL-friendly Qwen, our strategy yields further gains (12.2\% $\to$ 18.1\% on AIME 24). This indicates that optimizing for distributional clarity is a universally beneficial objective, helping even capable models to further refine their decision boundaries.

\begin{figure}[t]
    \centering
    \includegraphics[width=0.98\linewidth]{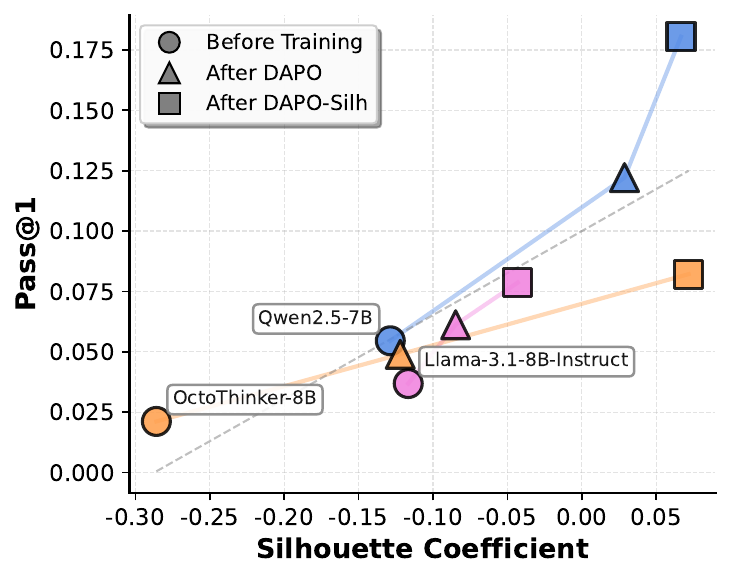}
    \vspace{-0.5em}
    \caption{Evolution of Silhouette Coefficient and Pass@1 across training stages on AIME24. The strong positive correlation ($r=0.815$) confirms that performance gains are driven by improved distributional clarity.}
    \label{fig:silhouette_evolution}
    \vspace{-1.2em}
\end{figure}

\noindent \textbf{Distributional Clarity Drives Performance.} Figure~\ref{fig:silhouette_evolution} tracks Silhouette Coefficient ($S$) and pass rate evolution across training, revealing strong correlation ($r=0.815$) that confirms the causal relationship between distributional clarity and performance. Model families exhibit distinct trajectories: less RL-friendly models (OctoThinker, Llama) start with extremely low $S$, where standard DAPO yields marginal clarity improvements while our strategy induces substantial $S$ increases corresponding to significant performance gains. Qwen starts with higher $S$ but still advances toward the high-performance region, demonstrating that distributional clarity remains a bottleneck even for capable models.

\subsection{Robustness and Ablation Analysis}
\label{sec:ablation_study}
Our analysis framework posits that the core driver of RL-Friendliness is distributional clarity—specifically, intra-class compactness and inter-class separation—rather than any particular choice of measurement metric or algorithm. To verify the generality of our approach, we conduct comprehensive ablation studies using the OctoThinker-8B backbone, as summarized in Table \ref{tab:ablation_generalization}.

\begin{table*}[t]
\centering
\small
\setlength{\tabcolsep}{4pt} 
\resizebox{0.95\textwidth}{!}{ 
\begin{tabular}{lccccccc} 
\toprule
\textbf{Method Setting} & \textbf{AIME24} & \textbf{AIME25} & \textbf{MATH500} & \textbf{AMC23} & \textbf{Minerva} & \textbf{Olympiad} & \textbf{Average} \\
\midrule
\multicolumn{8}{l}{\textit{\textbf{Default Setting}}} \\ 
DAPO & 4.9 & 2.1 & 59.4 & 46.9 & 27.6 & 25.8 & 27.8 \\
DAPO-Silhouette ($\beta=0.2$) & \textbf{8.2} & 5.0 & 62.6 & 47.2 & 29.6 & 27.8 & 30.1 \\
\midrule
\multicolumn{8}{l}{\textit{\textbf{Metric Generalization}}} \\
DAPO-Fisher Ratio & 7.5 & 4.5 & 63.6 & 44.4 & 29.0 & \textbf{29.9} & 29.8 \\
\midrule
\multicolumn{8}{l}{\textit{\textbf{Mechanism Validation}}} \\
Inter-class Separation Only & 5.8 & 2.2 & 62.4 & 48.0 & 26.7 & 27.6 & 28.8 \\
Pass-Rate Reweighting & 5.0 & 2.6 & 61.1 & \textbf{49.7} & 27.2 & 26.8 & 28.7 \\
Random Reweighting & 5.0 & 1.8 & 60.5 & 42.8 & 28.8 & 26.3 & 27.5 \\
\midrule
\multicolumn{8}{l}{\textit{\textbf{Hyperparameter Sensitivity}}} \\
DAPO-Silhouette ($\beta=0.1$) & 6.9 & 3.4 & 62.8 & 44.7 & 28.0 & 26.9 & 28.8 \\
DAPO-Silhouette ($\beta=0.5$) & 6.6 & \textbf{6.3} & \textbf{65.5} & 46.6 & \textbf{30.4} & 29.6 & \textbf{30.8} \\
\bottomrule
\end{tabular}
}
\vspace{-0.5em}
\caption{Robustness and ablation analysis on OctoThinker-8B. We confirm the effectiveness of our strategy across different metrics (Fisher Ratio) and varying hyperparameter settings ($\beta$). Mechanism Validation compares our full Silhouette approach against a variant optimizing only separation, outcome-based reweighting (Pass-Rate), and random noise.}
\label{tab:ablation_generalization}
\vspace{-1.2em}
\end{table*}

\noindent \textbf{Metric Generalization.}
We first investigate whether the performance gain is an artifact of the Silhouette Coefficient. We implement an alternative variant using the Fisher Ratio \citep{pattern}, a classical statistic that explicitly measures the ratio of inter-class variance to intra-class variance. By integrating this metric into DAPO via a similar reweighting strategy (details in Appendix~\ref{sec:appendix_fisher}), we observe performance improvements comparable to the Silhouette Coefficient. This confirms that the benefit stems from emphasizing distributional clarity itself, independent of the specific mathematical formulation used to quantify it.

\noindent \textbf{Mechanism Validation.}
To verify that distributional clarity specifically drives the gains, we evaluate three alternative strategies (Table~\ref{tab:ablation_generalization}). First, we test an Inter-class Separation Only variant, which modifies the weight to solely maximize the margin between correct and incorrect probabilities, ignoring intra-class compactness. While this yields improvements over the baseline (28.8\% vs. 27.8\%), it falls short of the full Silhouette strategy (30.1\%). This underscores that intra-class compactness is necessary for optimal RL-Friendliness. We further compare against Pass-Rate Reweighting and Random Reweighting. Pass-Rate Reweighting, which prioritizes low-success samples, achieves only 28.7\%, indicating that the benefits of our Silhouette strategy arise from distributional clarity rather than hard sample mining based on coarse outcome statistics. Random Reweighting yields 27.5\%, below the baseline, confirming that improvements require informative signals, not arbitrary reweighting. Collectively, these comparisons establish the necessity of Silhouette-Aware Reweighting for capturing signals that partial metrics and heuristic baselines fail to exploit.

\noindent \textbf{Hyperparameter Sensitivity.}
We analyze the impact of reweighting intensity $\beta$ by evaluating $\beta \in \{0.1, 0.2, 0.5\}$. Table \ref{tab:ablation_generalization} show that our method proves highly robust, with all $\beta$ values consistently outperforming the DAPO baseline. The aggressive setting ($\beta=0.5$) achieves the highest overall average (30.8\%), particularly excelling on MATH-500 and Minerva. However, the moderate setting ($\beta=0.2$) yields the best performance on the challenging AIME 2024 benchmark (8.2\% vs. 6.6\%). Consequently, we adopted $\beta=0.2$ as the default for our main experiments to balance performance across diverse difficulty benchmarks, while noting that higher $\beta$ values may offer further potential.

\section{Related Work}

\noindent \textbf{RLVR.}
RLVR has emerged as a dominant paradigm for enhancing reasoning capabilities~\citep{deepseekr1,deepseekmath_v2,kimik2,minimaxm1,deepseekv3_2}. Recent studies have explored reshaping reward or advantage by leveraging the model's intrinsic generation metrics to improve exploration efficiency. For instance, \citet{reasoning_with_exploration} incorporates token-level entropy into the advantage function, while others utilize model uncertainty—derived from perplexity or confidence scores—to refine credit assignment~\citep{cde_tencent,ucas_kuaishou,icpo_xiaomi,seedgrpo,entropymechanism}. Unlike these methods that focus on single-sample uncertainty, we reshape the advantage function using the group-level probability landscape to enforce distributional clarity.

\noindent \textbf{Data-Centric Approaches to Reasoning.}
Prior research largely attributes RL efficacy to data composition~\citep{spurious,midtraining_survey1,midtraining_survey2}, advocating for specific reasoning patterns like self-verification~\citep{cognitive} or high-quality mathematical mid-training~\citep{octothinker,reinforcementmidtraining}. Similarly, systematic studies highlight the benefits of front-loading reasoning data~\citep{front_loading} and bridging syntactic gaps~\citep{midtraining_bridges,interplay} to establish a robust foundation. Unlike these approaches that focus on \textit{what} models learn, we propose an orthogonal perspective: analyzing \textit{how} the intrinsic probability landscape affects trainability. We identify distributional clarity—specifically intra-class compactness and inter-class separation—as a prerequisite for effective RL optimization.

\section{Conclusion}

In this work, we investigated the disparity in RL-Friendliness across foundation models, proposing a shift from purely data-centric views to an intrinsic distributional perspective. Our three-stage analysis identified distributional clarity as a critical determinant of RL-Friendliness. we demonstrated that poor distributional clarity (quantified by the Silhouette Coefficient) is strongly associated with high-severity logic errors and reasoning instability. Validating this insight, our Silhouette-Aware Reweighting strategy significantly enhanced the performance of less RL-friendly models by prioritizing samples with ambiguous distributions. Our findings highlight that beyond data composition, the intrinsic distributional properties of foundation models serve as a fundamental prerequisite for effective reinforcement learning.

\section*{Limitations}

Our Silhouette-Aware Reweighting strategy relies on group-relative statistics derived from multiple rollouts (set to $G=16$) to estimate distributional clarity. While this statistical approach effectively stabilizes the training signal, the precision of the Silhouette Coefficient inherently benefits from a sufficient sample size. In scenarios with extremely restrictive sampling budgets or where generation diversity is severely collapsed, the distributional estimation may become less robust. However, it is worth noting that modern RLVR algorithms, such as GRPO and DAPO, already necessitate group sampling for baseline advantage estimation; thus, our approach leverages existing computational structures rather than introducing additional sampling overhead.

Additionally, our study focuses on models generating standard Chain-of-Thought reasoning paths, rather than the extremely long reasoning trajectories (e.g., exceeding 10k tokens) recently observed in some specialized reasoning models. We adopted this setting to facilitate a controlled comparison across diverse open-weight model families, such as Llama-3.1, which may not inherently support such extended contexts without specific adaptation. Furthermore, limiting the response length allows for extensive ablation studies within accessible computational resources, though investigating the distributional properties of these ultra-long reasoning processes remains a promising direction.

\bibliography{custom}

\appendix

\section{Details of Three-Stage Analysis}
\label{sec:appendix}

In this section, we provide supporting materials and detailed implementation settings for the three-stage analysis framework presented in Section \ref{sec:three_level}. This includes additional performance visualizations (Phenomenon Level) and the fine-grained taxonomies and prompts used for behavioral interpretation (Interpretation Level).

\subsection{Phenomenon Level: Additional Pass Rate Analysis}
\label{sec:appendix_pass_rate}

In Section \ref{sec:outcome_level}, we presented the per-problem pass rate comparison on the AIME 2024 benchmark. To further validate that the observed performance disparity is a consistent phenomenon across different mathematical reasoning tasks, we conduct the corresponding analysis on the MATH-500 benchmark.

\begin{figure}[h]
    \centering
    \includegraphics[width=0.98\linewidth]{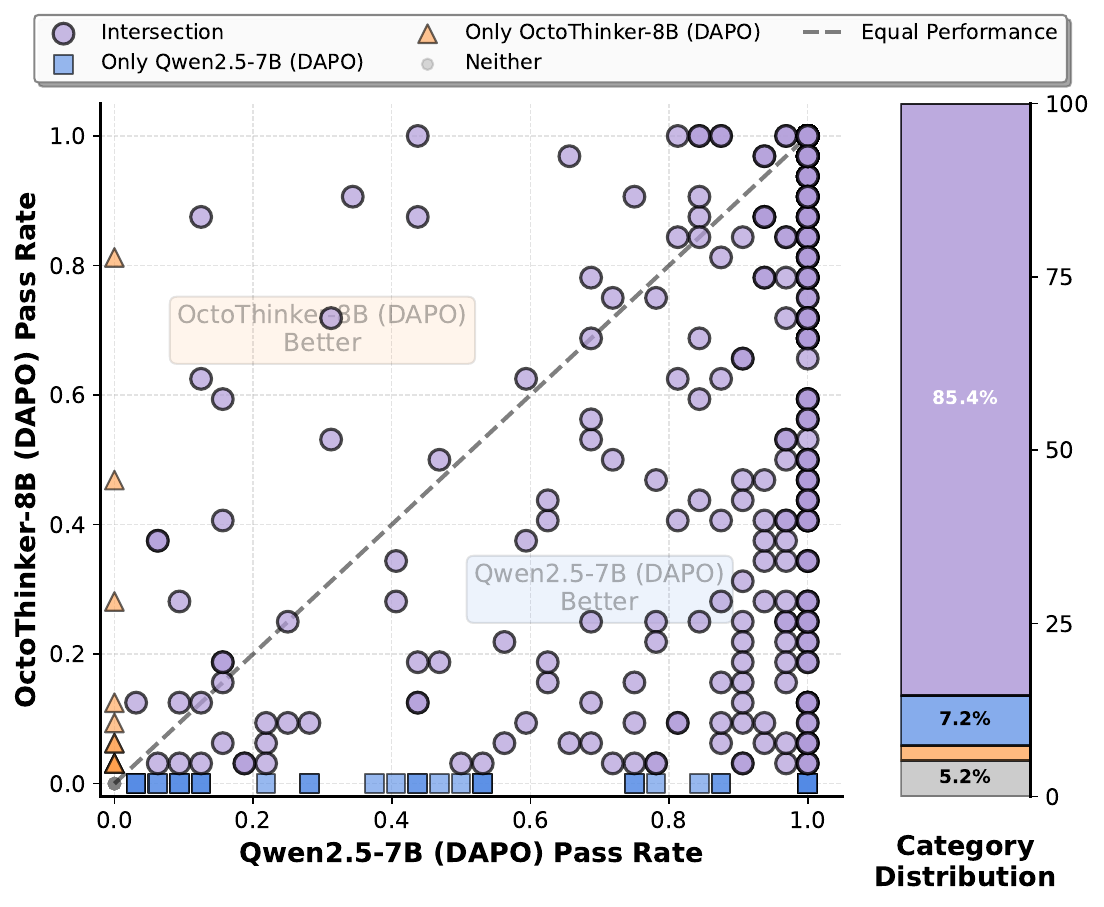}
    \caption{Per-problem pass rate comparison between Qwen2.5-7B (DAPO) and OctoThinker-8B (DAPO) on the MATH-500 benchmark. Despite a large intersection of mutually solvable problems, Qwen demonstrates higher pass rates on these shared queries compared to OctoThinker, mirroring the trend observed in AIME 2024.}
    \label{fig:pass_rate_comparison_math500}
    \vspace{-1.em}
\end{figure}

As shown in Figure \ref{fig:pass_rate_comparison_math500}, the results on MATH-500 exhibit a pattern highly consistent with AIME 2024. We observe a substantial intersection in the problem sets solvable by both models, indicating that they share comparable latent capabilities. However, within this large shared domain, Qwen2.5-7B achieves higher pass rates. This further confirms that RL-friendly models possess a superior capacity to assign high probability mass to correct solutions across diverse problem sets.

\subsection{Interpretation Level: Error Attribution Details}
\label{sec:appendix_error}

\noindent \textbf{Error Taxonomy.} 
We construct a comprehensive taxonomy to categorize incorrect responses based on the root cause of the failure. As shown in Table \ref{tab:error_taxonomy}, these categories are grouped into three severity levels: High (Fundamental), Mid (Execution), and Low (Presentation). This hierarchy allows us to distinguish between deep cognitive failures and superficial implementation mistakes.

\begin{table*}[htbp]
    \centering
    \small
    \renewcommand{\arraystretch}{1.4} 
    \begin{tabularx}{\textwidth}{@{}l l X@{}} 
    \toprule
    \textbf{Category} & \textbf{Code} & \textbf{Description} \\
    \midrule
    
    \multirow{6}{*}{\textbf{\makecell[l]{Fundamental\\Errors\\(High)}}} 
      & E1.1 & \textbf{Misunderstanding Question:} Misread variables, function definitions, or the goal. \\
      & E1.2 & \textbf{Constraint Violation:} Ignored constraints (e.g., ``positive integers'', ``distinct''). \\
      & E2.1 & \textbf{Knowledge Error:} Used wrong mathematical formulas, theorems, or facts. \\
      & E2.2 & \textbf{Planning/Method Error:} The chosen approach or model was fundamentally flawed. \\
      & E5.1 & \textbf{Repetition Loop:} Entered a degenerative loop repeating the \text{same} sequence. \\
      & E5.2 & \textbf{Irrelevant/Incoherent:} Failed to address the specific question or generated noise. \\
    \midrule
    
    \multirow{2}{*}{\textbf{\makecell[l]{Execution\\Errors\\(Mid)}}} 
      & E3.1 & \textbf{Calculation/Execution Error:} Correct formula but failed arithmetic or algebraic manipulation. \\
      & E3.2 & \textbf{Step Hallucination:} Invented intermediate values or steps without basis. \\
    \midrule
    
    \multirow{2}{*}{\textbf{\makecell[l]{Presentation\\Errors\\(Low)}}} 
      & E4.1 & \textbf{Format Error:} Correct answer but failed to format (e.g., missing \texttt{\textbackslash boxed\{\}}). \\
      & E4.2 & \textbf{Premature Stop:} Generation cut off before completion. \\

    \bottomrule
    \end{tabularx}
    \caption{Fine-grained taxonomy for error attribution. Errors are categorized by severity: High (Fundamental), Mid (Execution), and Low (Presentation).}
    \label{tab:error_taxonomy}
\end{table*}

\noindent \textbf{Evaluation Prompt.} 
We employ Qwen3-32B \citep{qwen3} as the judge model to classify errors. The prompt includes the full taxonomy definitions to ensure consistent attribution. The exact prompt template is provided in Figure \ref{fig:prompt_error_attribution}.

\begin{figure*}[htbp]
\centering
\begin{promptbox}[Prompt for Error Attribution Analysis]{blue_dist}
\small
You are a mathematical reasoning analyst. Your task is to analyze a MODEL RESPONSE to a math question that resulted in an INCORRECT answer. Identify the primary cause of the error based on the following taxonomy.\\
\\
\textbf{Taxonomy for Incorrect Outcomes:}\\
1. \textbf{[E1.1] Misunderstanding Question:} Misread variables, function definitions, or the goal.\\
2. \textbf{[E1.2] Constraint Violation:} Ignored constraints (e.g., ``positive integers'', ``distinct'', ``minimum'').\\
3. \textbf{[E2.1] Knowledge Error:} Used wrong mathematical formulas, theorems, or facts.\\
4. \textbf{[E2.2] Planning/Method Error:} The chosen approach/model was fundamentally flawed.\\
5. \textbf{[E3.1] Calculation/Execution Error:} Formula was correct, but arithmetic or algebraic manipulation failed.\\
6. \textbf{[E3.2] Step Hallucination:} Invented intermediate values or steps without basis.\\
7. \textbf{[E4.1] Format Error:} Answer calculated correctly but failed to format (e.g., missing \texttt{\textbackslash boxed\{\}}).\\
8. \textbf{[E4.2] Premature Stop:} Generation cut off before completion.\\
9. \textbf{[E5.1] Repetition Loop:} The model entered a degenerative loop, repeating the \text{same} phrase, number, or sequence endlessly.\\
10. \textbf{[E5.2] Irrelevant/Incoherent:} The response fails to address the specific question. It may be unrelated text, pure text completion, gibberish, or nonsensical noise.\\
11. \textbf{[E6.1] Other:} The response fits none of the above categories.\\
\\
\textbf{Output Format (JSON):}\\
\{\\
\indent \texttt{``category\_code''}: ``E3.1'',\\
\indent \texttt{``reason''}: ``Brief explanation of why this category was chosen.''\\
\}\\
\\
$[$Question$]$\\
\{question\}\\
\\
$[$Ground Truth$]$\\
\{ground\_truth\}\\
\\
$[$Model Response$]$\\
\{model\_response\}\\
\\
Analyze the response based on the Ground Truth and provide the attribution category code and reason in JSON.
\end{promptbox}
\caption{The LLM-as-a-judge prompt used for error attribution analysis.}
\label{fig:prompt_error_attribution}
\end{figure*}

\subsection{Interpretation Level: Solution Stability Details}
\label{sec:appendix_solution_stability}

In this subsection, we describe the methodology used to quantify the stability of reasoning paths generated by the models.

\noindent \textbf{Clustering Algorithm.}
To group correct solutions based on their semantic method, we employ an incremental clustering approach driven by an LLM judge. Unlike traditional embedding-based clustering which may fail to capture subtle logic differences, our method utilizes pairwise comparisons to determine if two solutions rely on the \text{same} core mathematical strategy. The detailed procedure is outlined in Algorithm \ref{alg:clustering}. For each problem, we maintain a list of distinct solution clusters. For every new correct response, we compare it against the representative solution of existing clusters. If a match is found, the response is assigned to that cluster; otherwise, it forms a new cluster.

\begin{algorithm*}[htbp] 
\caption{Incremental Solution Method Clustering}
\label{alg:clustering}
\begin{algorithmic}[1]
\Require Set of correct solutions $\mathcal{R} = \{o_1, o_2, \dots, o_N\}$ for a query $q$
\Require LLM Judge $\mathcal{J}(o_a, o_b) \to \{0, 1\}$
\Ensure Set of method clusters $\mathcal{C} = \{c_1, c_2, \dots, c_K\}$

\State Initialize clusters $\mathcal{C} \leftarrow \emptyset$
\For{each solution $o_i$ in $\mathcal{R}$}
    \State $assigned \leftarrow \text{FALSE}$
    \For{each cluster $c_k$ in $\mathcal{C}$}
        \State $o_{rep} \leftarrow \text{Representative}(c_k)$ \Comment{First element of cluster}
        \If{$\mathcal{J}(o_i, o_{rep}) == 1$}
            \State Add $o_i$ to $c_k$
            \State $assigned \leftarrow \text{TRUE}$
            \State \textbf{break}
        \EndIf
    \EndFor
    \If{$assigned == \text{FALSE}$}
        \State Create new cluster $c_{new} = \{o_i\}$
        \State Add $c_{new}$ to $\mathcal{C}$
    \EndIf
\EndFor
\State \Return $\mathcal{C}$
\end{algorithmic}
\end{algorithm*}

\noindent \textbf{Evaluation Prompt.}
The LLM judge determines whether two solutions share the \text{same} fundamental approach. The specific criteria for \text{same} method (e.g., identical theorems or logical strategies) versus different method (e.g., geometric vs. algebraic approaches) are explicitly defined in the system prompt to ensuring consistency. The complete prompt used for this pairwise comparison is presented in Figure \ref{fig:prompt_method_judge}.

\section{Gradient Variance Analysis}
\label{sec:gradient_variance}
For a query $q$ with correct responses sampled from $\pi^+(\cdot|q)$ and incorrect responses from $\pi^-(\cdot|q)$, the GRPO policy gradient can be written as \citep{2-GRPO}:
\begin{equation}
\begin{split}
\nabla_\theta J &\propto \mathbb{E}_{o_j \sim \pi^+} [\nabla_\theta \pi_\theta(o_j|q)] \\
&\quad - \mathbb{E}_{o_k \sim \pi^-} [\nabla_\theta \pi_\theta(o_k|q)]
\end{split}
\end{equation}

Define $g_+ = \mathbb{E}_{o_j \sim \pi^+} [\nabla_\theta \pi_\theta(o_j|q)]$ and $g_- = \mathbb{E}_{o_k \sim \pi^-} [\nabla_\theta \pi_\theta(o_k|q)]$ as the expected gradients over correct and incorrect responses. Assuming that correct and incorrect responses are sampled independently, the variance of the policy gradient is:
\begin{equation}
\text{Var}(\nabla_\theta J) = \text{Var}(g_+) + \text{Var}(g_-)
\end{equation}

For a language model with softmax output layer, the derivative of the probability $\pi$ with respect to the logit $z$ is $\frac{\partial \pi}{\partial z} = \pi(1 - \pi)$. By the chain rule, the gradient of the probability with respect to model parameters $\theta$ can be written as:
\begin{equation}
\nabla_\theta \pi_\theta(y|x) = \pi(y|q)(1 - \pi(y|q)) \cdot \nabla_\theta z
\end{equation}

Since hidden representations in transformers are normalized and exhibit stable variance, the stochasticity of gradient updates is predominantly governed by the $\pi(1-\pi)$ term. We obtain:
\begin{equation}
\text{Var}(\nabla_\theta J) \propto \text{Var}_{\pi^+}[\pi(1-\pi)] + \text{Var}_{\pi^-}[\pi(1-\pi)]
\end{equation}
where we use the shorthand $\text{Var}_{\pi^+}[\pi(1-\pi)]$ to denote $\text{Var}_{o_j \sim \pi^+}[\pi(o_j|q)(1-\pi(o_j|q))]$. When probabilities of correct responses are tightly clustered around some value $\pi_+$, the variance $\text{Var}_{o_j \sim \pi^+}[\pi(1-\pi)]$ is small. Similarly, when incorrect responses cluster tightly around $\pi_-$, the variance $\text{Var}_{o_k \sim \pi^-}[\pi(1-\pi)]$ is small.

\section{Experimental Setup Details}
\label{sec:appendix_experimental_setup}

In this section, we provide the detailed configurations for both the training and evaluation phases. 

\subsection{Training Configuration}
\label{sec:appendix_training_config}

We implement our training pipeline using the verl framework \citep{verl}. All models are trained on a node of 8 NVIDIA H800 GPUs. The training utilizes the DAPO algorithm with the hyperparameters detailed in Table \ref{tab:hyperparameters}.

\begin{table}[htbp]
    \small
    \centering
    \resizebox{0.8\columnwidth}{!}{%
    \begin{tabular}{lc}
        \toprule
        \textbf{Hyperparameter} & \textbf{Value} \\
        \midrule
        \multicolumn{2}{l}{\textit{\textbf{Data Configuration}}} \\
        Max prompt length & 2048 \\
        Max response length & 8192 \\
        \midrule
        \multicolumn{2}{l}{\textit{\textbf{DAPO Algorithm Configuration}}} \\
        Advantage estimator & GRPO \\
        Clip ratio (low) & 0.2 \\
        Clip ratio (high) & 0.28 \\
        Responses per prompt & 16 \\
        Sampling temperature & 1.0 \\
        Sampling top-p & 1.0 \\
        KL in reward & False \\
        KL loss & False \\
        \midrule
        \multicolumn{2}{l}{\textit{\textbf{Optimization Configuration}}} \\
        Optimizer & AdamW \\
        Learning rate & 1e-6 \\
        Learning rate warmup steps & 10 \\
        Weight decay & 0.1 \\
        Gradient clipping & 1.0 \\
        Batch size & 512 \\
        Mini-batch size & 32 \\
        Total training steps & 200 \\
        \bottomrule
    \end{tabular}%
    }
    \caption{Training hyperparameters.}
    \label{tab:hyperparameters}
\end{table}

\subsection{Evaluation Configuration}
\label{sec:appendix_eval_config}

During evaluation, we generate multiple reasoning paths for each query. The decoding parameters are set to \texttt{temperature=0.6} and \texttt{top\_p=0.95}, with a maximum token limit of 8192.

The number of sampled rollouts ($K$) varies by benchmark:
\begin{itemize}[leftmargin=13pt, topsep=0pt, itemsep=0pt]
    \item \textbf{AIME 2024, AIME 2025:} $K=256$.
    \item \textbf{MATH-500:} $K=32$.
    \item \textbf{AMC 23, Minerva, OlympiadBench:} $K=16$.
\end{itemize}
The final performance is reported as the average value calculated across these $K$ samples.

\subsection{Fisher Ratio Implementation}
\label{sec:appendix_fisher}

To verify the robustness of our approach across different distributional metrics, we implement a variant based on the Fisher Ratio ($F$). While the Silhouette Coefficient focuses on geometric cluster separation, the Fisher Ratio statistically quantifies the separation by comparing the squared difference of means to the sum of variances.

For a given query, let $\mu_{pos}, \sigma^2_{pos}$ and $\mu_{neg}, \sigma^2_{neg}$ denote the mean and variance of the sequence-level probabilities for correct and incorrect responses, respectively. The standard Fisher Ratio $F$ is calculated as:
\begin{equation}
    F = \frac{(\mu_{pos} - \mu_{neg})^2}{\sigma_{pos}^2 + \sigma_{neg}^2}
\end{equation}

Similar to our Silhouette strategy, we must handle cases where the distribution is inverted (i.e., incorrect responses have higher probabilities than correct ones). We define a rectified Fisher score $F'$:
\begin{equation}
    F' = 
    \begin{cases} 
    F & \text{if } \mu_{pos} \ge \mu_{neg} \\
    -F & \text{if } \mu_{pos} < \mu_{neg}
    \end{cases}
\end{equation}

The reweighting factor $w(q)$ is then computed using an exponential decay function. Unlike the Silhouette Coefficient which is bounded in $[-1, 1]$, the Fisher Ratio can vary widely in magnitude. Therefore, we apply a clamping mechanism to ensure training stability:
\begin{equation}
    w(q) = \text{clip}\left(\exp(-\beta \cdot F'), 0.95, 1.05\right)
\end{equation}
where $\text{clip}(x, \text{min}, \text{max})$ restricts the weight within the specified range. For our experiments, we set the sensitivity hyperparameter $\beta = 0.01$. Finally, the advantage is modulated as $\hat{A}_{new} = \hat{A}_{old} \cdot w(q)$.

\begin{figure*}[htbp]
\centering
\begin{promptbox}[Prompt for Solution Method Comparison]{blue_dist}
\small
You are a mathematics expert and logic analyst.
Your task is to compare two CORRECT solutions to the \text{same} math problem and determine if they use the \textbf{\text{same} core mathematical method/approach}.\\
\\
\textbf{Criteria for \text{same} Method:}
\begin{itemize}
    \setlength\itemsep{0pt} \setlength\parskip{0pt} \setlength\parsep{0pt} 
    \item They use the \text{same} key theorems, formulas, or logical strategy (e.g., both use ``Coordinate Geometry'' or both use ``Mass Point Geometry'').
    \item Ignore differences in variable names, calculation order, or verbosity.
    \item Ignore differences in how the answer is formatted if the derivation logic is identical.
\end{itemize}
\textbf{Criteria for Different Method:}
\begin{itemize}
    \setlength\itemsep{0pt} \setlength\parskip{0pt} \setlength\parsep{0pt}
    \item One uses a geometric approach while the other uses an algebraic/trigonometric approach.
    \item One uses a brute-force enumeration while the other uses a combinatorial formula.
    \item One uses a specific theorem (e.g., Fermat's Little Theorem) while the other uses pattern finding.
\end{itemize}
\textbf{Output JSON Format:}\\
\{\\
\indent \texttt{``is\_\text{same}\_method''}: \textit{boolean},\\
\indent \texttt{``reason''}: ``Brief explanation of the similarity or difference.''\\
\}\\
\\
$[$Question$]$\\
\{question\}\\
\\
$[$Solution A$]$\\
\{solution\_a\}\\
\\
$[$Solution B$]$\\
\{solution\_b\}\\
\\
Do Solution A and Solution B use the fundamentally \text{same} mathematical method?
\end{promptbox}
\caption{The LLM-as-a-judge prompt used to determine if two solutions utilize the \text{same} reasoning method.}
\label{fig:prompt_method_judge}
\end{figure*}

\end{document}